\pdfoutput=1

\documentclass{article}

\usepackage[final,nonatbib]{nips_2016}

\usepackage[
bookmarks=true,
colorlinks=true,
linkcolor=red,
urlcolor=blue,
citecolor=blue,
pdftex,
bookmarks=true,
linktocpage=true, 
hyperindex=true
]{hyperref}

\usepackage{hyperref}       

\hypersetup{
backref=true
bookmarksnumbered,
pdfstartview={FitH},
citecolor={blue},
linkcolor={red},
urlcolor={black},
pdfpagemode={UseOutlines}
}
\usepackage[utf8]{inputenc} 
\usepackage[T1]{fontenc}    
\usepackage{hyperref}       
\usepackage{url}            
\usepackage{booktabs}       
\usepackage{amsfonts}       
\usepackage{nicefrac}       
\usepackage{microtype}      
\usepackage{amsmath,amssymb,epsfig,amsthm,times,ifthen}
\usepackage{graphicx}
\usepackage{epsf,epic,psfrag,mathabx}
\usepackage{enumerate}
\usepackage{multirow}
\usepackage[center]{caption}
\usepackage[table]{xcolor}

\DeclareMathOperator{\vect}{vec}

\newtheorem{theorem}              {Theorem}     [section]

\theoremstyle{remark}
\newtheorem{remark}     {Remark}

\theoremstyle{definition}         

\newtheorem{example}     {Example}

\newcommand{\sym}[1]{\mbox{\textit{sym}}(#1)}

\newcommand{\remove}[1]{}
\newcommand{\supplementaMaterial}[1]{#1}

\newcommand{\TNkernels}{Symmetric kernels }
\newcommand{\tnkernels}{symmetric kernels }
\newcommand{\tnkernel}{symmetric kernel }
\newcommand{\tOne}{\cellcolor{green!10}}
\newcommand{\tTwo}{\cellcolor{red!20}}
\newcommand{\tThr}{\cellcolor{blue!20}}
\newcommand{\tFou}{\cellcolor{yellow!20}}

\title{Structured Convolution Matrices for Energy-efficient Deep learning}

\author{
  Rathinakumar Appuswamy \\
  IBM Research--Almaden\\
  \texttt{rappusw@us.ibm.com} \\
  \And
  Tapan K.~Nayak \\
    IBM Research--Almaden\\
  \texttt{tknayak@us.ibm.com} \\
  \And
  John Arthur \\
    IBM Research--Almaden\\
  \texttt{arthurjo@us.ibm.com} \\
  \And
  Steven Esser\\
  \texttt{sesser@us.ibm.com} \\
    IBM Research--Almaden\\
  \And
  Paul A.~Merolla\\
  \texttt{pameroll@us.ibm.com} \\
    IBM Research--Almaden\\
  \And
  Jeffrey L.~Mckinstry\\
    IBM Research--Almaden\\
  \texttt{jlmckins@us.ibm.com} \\
  \And
  Timothy Melano\\
    IBM Research--Almaden\\
  \texttt{tmelano@us.ibm.com} \\
  \And
  Myron Flickner\\
    IBM Research--Almaden\\
  \texttt{mdflickner@us.ibm.com} \\
  \And
  Dharmendra S.~Modha\\
    IBM Research--Almaden\\
  \texttt{dmodha@us.ibm.com} \\
  }

\begin{document}
\maketitle

\begin{abstract}
We derive a relationship between network representation in  energy-efficient neuromorphic architectures and block Toplitz convolutional matrices. 
Inspired by this connection, we develop deep convolutional networks using a family of structured convolutional matrices and achieve state-of-the-art trade-off between 
energy efficiency and classification accuracy for well-known image recognition tasks. 
We also put forward  a novel method to train binary convolutional networks by utilising an existing connection between 
noisy-rectified linear units and binary activations.
\end{abstract}

\section{Introduction}
Deep convolutional networks have, in recent times, achieved near-human performance on an array of visual, auditory, and other cognitive tasks \cite{alexNet,silver2016mastering}.
The intriguing possibility of delivering deep learning applications on mobile devices, as well as providing energy-efficient cognitive solutions on the cloud have inspired 
an increasing number of researchers to search for low-precision state-of-the-art convolutional networks that can be deployed on extremely energy-efficient 
platforms \cite{Steve,binaryNet,binaryNet2,binaryConnect,compressMobile,XNOR}.

Binary convolutional networks that use binary convolutional kernels, and binary neuron activations are ideally suited to be run on low-power neuromorphic architectures that use 
spike-based communication \cite{Merolla}. 
In addition, storage and computational efficiency may be gained by using structured matrices in convolutional layers.
Using structured matrices in the fully connected layers (also known as linear layers) of deep networks has been studied in the literature \cite{smallFootprint,Moczulski2015,ionescu2015matrix} with the objective of reducing the number of learned parameters. The main idea behind these approaches is to restrict the connectivity matrix of a (fully connected) layer to be from a known 
family of matrices, which are parametrised by a few variables and adapt the backpropagation to update those variables.
On the other hand, reducing the memory and computation requirements of convolutional layers has been addressed by several approaches to model compression \cite{rankConstrainedTop,vecQuant,deepComp,compressMobile,hashNet,optimalBrain,bottleneckFeaturesTara,bottleneckFeaturesPetr,lowRankMatrixTara,EIE} -- mostly after training.

In this work, we propose the use of structured matrices in convolutional layers that are inspired by low-power neuromorphic hardware architectures \cite{Merolla,spinnaker,pfeil2012six,schmuker2014neuromorphic,moradi2014event,park201465k,EIE}.
By connecting the efficient weight representation schema used in neuromorphic architectures  \cite{Merolla,EIE} with block Toeplitz matrices that arise in discrete convolution, 
we identify a family of convolution kernels that are naturally hardware efficient. 
The primary motivation behind our investigation 
is in the tradition of discovering algorithms that are native to a chosen architecture \cite{frigo2005design,maass2004computational,dongarra2000guest,hoare1962quicksort} and thereby harvesting the best that a particular architecture offers.

We incorporate learning structured convolutional matrices into traditional stochastic gradient descent \cite{bottou2010large} so that the trained inference networks are hardware-ready.
Furthermore, we exploit a known equivalence between stochastic bounded rectified linear units and deterministic threshold neurons and propose a novel approach to training networks with binary neuron activations. This procedure allows us to obtain  accurate gradient estimates during backward step, and speed-up convergence of training and enriches the library of tools available to train low-precision deep neural networks. We evaluate our system on Cifar10 data set and compare against best energy vs accuracy numbers reported on currently available hardware \cite{Steve} -- our approach reduces the number of TrueNorth cores required to achieve $87.5$\% on Cifar10 from $31872$ TrueNoth cores in \cite{Steve} to $13216$ cores.

We begin the next section by discussing binary convolutional networks and their suitability for neuromorphic hardware and introduce the weight representation mechanism used in TrueNorth architecture \cite{Merolla}. In Section~\ref{Sec:tnk}, we discuss the structure of discrete convolution matrices and present our main result connecting that structure 
with the weight representation in TrueNorth -- thereby identifying the family of structured convolution matrices that efficiently map to that architecture.
Section~\ref{Sec:train} outlines our methodology for training networks to use only structured convolution kernels, and discusses how the connection between noisy ReLUs and binary neurons is exploited during training.
We summarize our experimental results in Section~\ref{Sec:res} followed by concluding remarks in Section~\ref{Sec:con}.

\section{Sparse Binary convolutional networks}

Binary convolutional networks \cite{binaryNet2,Steve} are a particularly elegant instance of low-precision  computing targeted toward object recognition tasks.
Motivated by the possibility of deploying state-of-the-art image recognition convolutional networks on low-power mobile devices, there has been an increasing interest around 
binary networks \cite{binary1,binaryNet,binaryNet2,XNOR,binaryConnect,Steve}.
Such networks use $\{-1,1\}$-valued (or, $\{0,1\}$) neurons throughout the network and thus best suited to be deployed on neuromorphic hardware using spiking neurons.
Although we will restrict our focus to TrueNorth architecture in this study \cite{Merolla}, most of our conclusions are also applicable to other platforms offering similar features \cite{spinnaker,pfeil2012six,schmuker2014neuromorphic,moradi2014event,park201465k,EIE}.

A TrueNorth chip consists of a network of neurosynaptic  cores  with  programmable  connectivity,  synapses,  and
neuron parameters \cite{Merolla}.  
It is a multicore array where each core consists of
$256$ input  lines, $256$ neurons,  and  a $256 \times 256$
synaptic  crossbar  array.   
Each input line in a core can connect to one neuron on any 
core through a spike router, and that input line is accessible to all of the $256$ neurons on that core
 through  the  crossbar,  thereby  resulting  in  block-wise
connectivity.
 All communication to-, from-, and within- chip is performed using spikes.
TrueNorth  neurons  use  a  variant  of  an  integrate-and-fire
model with 23 configurable parameters \cite{neuron} where a neuron's
state variable updates each tick (typically at 1000 ticks per second, though higher rates are possible).
Synapses  have  individually  configurable  on/off  states  and
have a strength assigned by look-up table.  Specifically, each
neuron has a 4-entry table of $8$-bit signed-integers,  
each  input  line  to  a  core  is  assigned an input-type of 1,  2,  3 or 4,  and each synapse then
determines its strength by using the input-type on its source
side to index into the table of the neuron on its target side.
In summary, for non-negative integers $L,N \le 256$,  the crossbar weight matrix is factored into three components:  
a) a $L \times N$ binary ($\{0,1\}$-valued) connectivity matrix C, (b) an input-type vector $g$ of length $L$ over the integers $\{1,2,3,4\}$, and (c) a set of strength-value functions $s_j : \{1,2,3,4\} \rightarrow \{-255,\ldots,255\}, j=1,2,\ldots,N$ -- one for each of the $N$ neurons on a core.
If we extend functions $s_j$ to operate on vectors element-wise, then
the  $L \times N$ weight matrix $M$ of a core can be written as 

\begin{align} \label{Eq:TNweightMatrix}
M(g,C,\{s_i\}_{i=1}^N) = 
\begin{bmatrix}
 s_1(g) & s_2(g) & \cdots & s_{N}(g)
\end{bmatrix} \circ C
\end{align}
where $\circ$ denotes the Hadamard product\footnote{Which is simply the element-wise product between matrices} between matrices. 


To account for the fact that a TrueNorth neuron can connect to at most $256$ input features,
as in \cite{Steve}, network structure in this paper constrained by partitioning features in each layer into
one  or  more  equally  sized  groups.
Neurons in a group connect to all the input features in that group, and output features from different groups are generated from disjoint set of input features\footnote{This is in contrast to typical convolutional network layer whose output features depend on the entire set of input features.}.   
The number of groups in a layer is chosen such that the total filter size (rows $\times$ columns $\times$ features)  
as well as the number of output features of  each  group  is  less  than  or  equal  to  $256$.
To further promote sparsity in connections, we allow the convolutional filters to be $\{-1,0,1\}$-valued.
The number of neurons that spike in response to an input image
influence the energy utilization as well as the number of classification per unit time that the underlying hardware
can support. With this in mind, we strive to train networks with sparse binary activations. 
All the networks studied in this work use $\{0,1\}$-valued neurons where average fraction of $1$'s is less than $20\%$.
All of the above considerations result in networks that are sparse in connectivity, weights, and activations. 

\section{\TNkernels} \label{Sec:tnk}

When discrete convolution is written as a matrix multiplication, the resulting `convolution matrix' has a block Toeplitz structure\footnote{We allow the Toeplitz matrices to be non-square.} \cite{gray2006toeplitz}. 
By establishing the connection between the weight representation in \eqref{Eq:TNweightMatrix} and Toeplitz matrices, we identify a set of convolution kernels ({\it \tnkernels\footnote{The name \textit{\tnkernels} is motivated by the fact that these kernels are generated using commutative members of the Symmetric group $S_4$--the group of all permutations on a set of four elements.}})
that efficiently map to the TrueNorth architecture.
We need a few more notations before giving a formal statement of this connection. 

For an arbitrary matrix $H$, we denote its $i$-th column by $H_i$ and its $(i,j)$-th element by $H_{i,j}$.
If $h$ is an arbitrary scalar-valued function on real numbers, we extend it to operate on matrices by applying it element-wise.
For two functions $h_1$ and $h_2$, $h_1 \cdot h_2$ denotes function composition.
Let $X \coAsterisk K$ denote the $2$-D convolution\footnote{For notational simplicity, we ignore the boundary effects and assume $n \ge 2l$ and convolution stride of $1$ throughout this section. The results presented here can be generalized to other cases.} between an $n \times n$ data matrix $X$ and an $l \times l$ convolution kernel matrix $K$. 
For any matrix $Y$, let $\vect(Y)$ denote the vectorization (i.e., concatenation of columns) of $Y$. \\
\textit{Fact 1 (See for example, \cite{gray2006toeplitz}):} If  $W$ denotes the convolution matrix such that
\begin{align} \label{Eq:vecConv}
\vect(X \coAsterisk K)^t = \vect(X)^t \, W
\end{align}
then $W$ is a block Toeplitz matrix of the form
\begin{footnotesize}
\begin{align} \label{Eq:Toeplitz}
W = 
\begin{bmatrix}
W_1 & 0  &  \cdots & 0 & 0  \\
W_2 & W_1 &  \cdots & 0 & 0 \\
\vdots & \vdots  & \cdots & \vdots &  \vdots \\
W_l & W_{l-1} & \cdots & 0 & 0\\
0 & W_l & \cdots & 0 & 0\\
\vdots & \vdots & \cdots & \vdots & \vdots \\
0 & 0 & \cdots & W_{1}  & 0 \\
0 & 0 & \cdots & W_2 & W_1 \\
\vdots & \vdots & \cdots & \vdots & \vdots \\
0 & 0 & \cdots & W_{l}  & W_{l-1} \\
0 & 0 & \cdots & 0 & W_l
\end{bmatrix}, \quad  \mbox{where submatrix} \; W_i = 
\begin{bmatrix}
K_{1,i} & 0 & \cdots & 0 & 0 \\
K_{2,i} & K_{1,i} & \cdots & 0 & 0 \\
\vdots  & \vdots & \cdots & \vdots & \vdots \\
K_{l,i} & K_{l-1,i} & \cdots & 0 & 0 \\
0 & K_{l,i} & \cdots & 0 & 0 \\
\vdots  & \vdots & \cdots & \vdots & \vdots \\
0 & 0 & \cdots & K_{1,i} & 0 \\
0 & 0 & \cdots & K_{2,i} & K_{1,i} \\
\vdots  & \vdots & \cdots & \vdots & \vdots \\
0 & 0 & \cdots & K_{l,i} & K_{l-1,i} \\
0 & 0 & \cdots & 0 & K_{l,i} \\
\end{bmatrix}\end{align}
\end{footnotesize}
 $0$ denotes the zero matrices of appropriate dimension, and for $i =1,2,\ldots,l$,  $W_i$ is an $n \times (n-l+1)$ Toeplitz matrix constructed from the $i$-th column of the convolution kernel $K$.
 Such an $n^2 \times n(n-l+1)$ matrix $W$  will be referred to as the {\it block Toeplitz convolution matrix for $K$}, and we write $W(K)$ to make its dependence on $K$ explicit.
The sparsity and circulant nature of convolution matrices are fully exploited in most approaches to accelerate training and inference \cite{van2014optimizing,cuDNN,qadeer2013convolution}. 

Suppose that the dimension of the data matrix $X$ is such that $n^2 \le 256$. 
Ideally, for every choice of kernel $K$ the corresponding convolution matrix $W(K)$ can be represented in the form of \eqref{Eq:TNweightMatrix} for some choice of 
type vector $g$, a set of strength functions, and a binary connectivity matrix $C$.
However, the representation in \eqref{Eq:TNweightMatrix} requires the input-type vector $g$ to be common to all the columns of the weight matrix. 
Hence, it is conceivable that not every block Toeplitz convolution matrix $W(K)$ can be represented in that form.
\begin{figure}[hht] 
\begin{center}
\scalebox{0.8}{\includegraphics{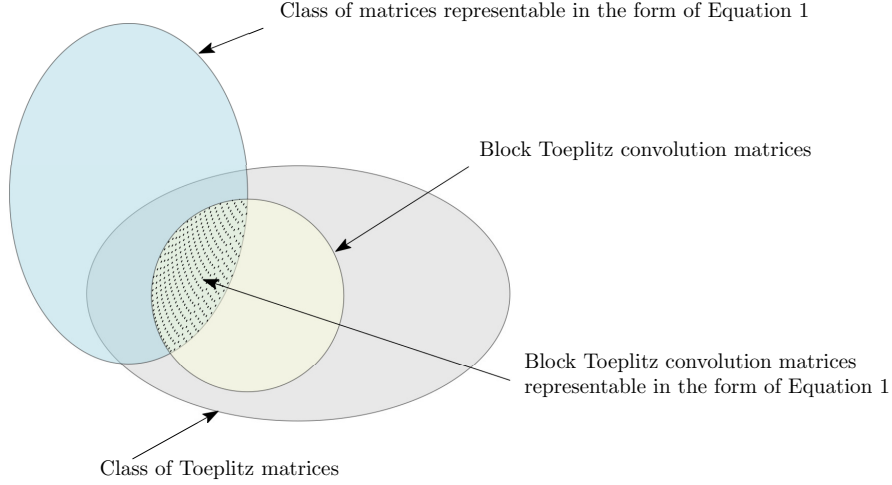}}
\end{center}
\caption{Venn diagram representation of relevant matrix families}
\label{Fig:matrixSpace}
\end{figure}
Figure~\ref{Fig:matrixSpace} provides a visual summary of the main result of this section: 
There is an one-to-one relationship between the set of convolution kernels and the set of block Toeplitz matrices of the form in \eqref{Eq:Toeplitz}.
Theorem~\ref{Th:toeplitz} asserts that only the set of block Toeplitz matrices associated with \tnkernels characterized by the theorem can be represented by \eqref{Eq:TNweightMatrix}. The characterization also suggests a constructive procedure to generate all symmetric kernels. The proof of the theorem is outlined in the appendix.
Example~\ref{Ex:toyExample}, also included in the appendix, explains 
the problem of relating the weight representation in \eqref{Eq:TNweightMatrix} and convolution matrices in more detail.
\begin{theorem} \label{Th:toeplitz}
Let $n$ and $l$ be non-negative integers such that $L = n^2 \le 256$, $N = n-l+1 \le 256$ and let $K$ be an $l \times l$ kernel matrix.
 There exist $ g, C$, and $s_1,\ldots,s_{N}$ as in \eqref{Eq:TNweightMatrix} such that 
$M(g,C,\{s_i\}_{i=1}^N) = W(K)$ if  $K$ satisfies 
\begin{align} \label{Eq:TNkernels}
K_{i,j} = B_{i,j} f(\sigma_1^{i-1} ( \sigma_2^{j-1} (\rho))) 
\end{align}
for some choice of 
\begin{enumerate}
\item commutative permutations $\sigma_1$, and $\sigma_2$ on $\{1,2,3,4\}$
\item seed element $\rho \in \{1,2,3,4\}$
\item function $f : \{1,2,3,4\} \rightarrow \{-255,\ldots,255\}$ 
\item $l \times l$  binary matrix $B$.
\end{enumerate}
If $K$ contains at least four distinct entries and no zeros, then the conditions on $K$ are also necessary.
\end{theorem}
We refer to the convolution kernels identified in Theorem~\ref{Th:toeplitz} as {\it \tnkernels} and use the notation $\sym{f,\rho,\sigma_1,\sigma_2,B}$ to refer to a particular instance.
The following two examples use the ordered-pair notation to define the functions $\sigma_1$,$\sigma_2$, and $f$ and show how some of the well-known convolutional operators are in the family of  \tnkernels identified by Theorem~\ref{Th:toeplitz}.
\begin{example}\label{Ex:1}
If $\rho = 1$, 
$\sigma_1 : \{(1\mapsto2),(2 \mapsto 1),(4 \mapsto 3),(3 \mapsto 4)\}$, 
$\sigma_2 : \{(1 \mapsto 2),(2 \mapsto 1),(4 \mapsto 3),(3 \mapsto 4)\}$, 
$f :  \{(1 \mapsto 4),(2 \mapsto -1),(3 \mapsto 4),(4 \mapsto 4)\}$, and 
\begin{align*}
B = \begin{bmatrix}
0 & 1 & 0 \\
1 & 1 & 1 \\
0 & 1 & 0 
\end{bmatrix}, \;  \mbox{then} \; K = \begin{bmatrix}
 0 & f(\sigma_2(\rho)) & 0\\
 f(\sigma_1(\rho)) & f(\sigma_1(\sigma_2(\rho))) & f(\sigma_1(\sigma_2^2(\rho))) \\
 0 & f(\sigma_1^2(\sigma_2(\rho))) & 0 
\end{bmatrix} =
\begin{bmatrix} 
0 & -1 & 0\\
-1 & 4 & -1\\
0  & -1 & 0 
\end{bmatrix}
\end{align*}
which is a widely used approximation to the spatial Laplacian operator.
\end{example}
\begin{example}
If $\rho = 1$,
$\sigma_1 : \{(1 \mapsto 1),(2 \mapsto 2),(3 \mapsto 3),(4 \mapsto 4)\}$,
$\sigma_2 : \{(1 \mapsto 2),(2 \mapsto 3),(3 \mapsto 4),(4 \mapsto 1)\}$,
$f : \{(1 \mapsto -1),(2  \mapsto -1),(3 \mapsto 1),(4 \mapsto 1)\}$, and 
\begin{align*}
B = \begin{bmatrix}
1 & 0 & 1\\
1 & 0 & 1\\
1 & 0 & 1
\end{bmatrix}, \quad \mbox{then} \quad K = \begin{bmatrix}
-1 & 0 & 1\\
-1 & 0 & 1\\
-1 & 0 & 1
\end{bmatrix}.
\end{align*}
$K$ is the vertical Prewitt operator and the corresponding horizontal operator also may be constructed similarly.
\end{example}

Notice that the set of \tnkernels include instances from the set of separable (e.g. Prewitt), as well as nonseparable (e.g. Laplacian) matrices.
A simple search reveals that there are $120$ unique pairs $\sigma_1$ and $\sigma_2$ of commutative permutations on the set of four elements that can be used 
in Equation~\ref{Eq:TNkernels}. Moreover, since we only consider $\{-1,0,1\}$-valued convolution kernels in the work, the function $f$ in \eqref{Eq:TNkernels} is restricted to be
$\{-1,1\}$-valued, thus there are $16$ possible choices for $f$. Since there are $4$ possible choices for $\rho$ and the components of $B$ are independent, there are a total of $2^{l^2} \times 16 \times 120 \times 4$ possible \tnkernels of dimension $l \times l$. 

We have limited the discussion thus far in this section to $2$-D convolutional matrices for ease of presentation.
However,  recall that deep convolutional networks use $3$-D matrices as convolutional kernels and we will now extend our definition of \tnkernels to cover this important case. An $l \times l \times m$ \textit{\tnkernel} is defined by
\begin{enumerate}
\item a pair of commutative permutations $\sigma_1$, and $\sigma_2$
\item an $m$-length seed vector $\rho \in \{1,2,3,4\}^m$
\item a function $f$ whose domain is $\{1,2,3,4\}$
\item a binary connectivity matrix $B$
\end{enumerate}
as 
\begin{align} \label{Eq:TNkernels3D}
K_{i,j,k} = B_{i,j,k} f(\sigma_1^{i-1} \sigma_2^{j-1} \rho_k).
\end{align}
A straight-forward counting argument as before reveals that there are $2^{m \, l^2} \times 16 \times 120 \times 4^m$ \tnkernels of dimension $l \times l \times m$.
Suppose that $l = 3$, and $m = 8$, there are about $10^{30}$  $\{-1,0,1\}$-valued kernels to choose from!

\section{Training} \label{Sec:train}
\subsection{Weights}
In a typical deep learning setting, the goal is to learn a set of kernels for every convolutional layer in order to minimize some loss function using a gradient-based optimization method.
In our case, we want the kernels to be from the class of symmetric kernels that efficiently map to TrueNorth architecture.  
Our strategy is as follows: At first, we allow every kernel $K$ in the network to be learned without any constraints.  
Next, because our goal is to generate a trained network that maps to TrueNorth, 
once the performance on the validation data saturates, the unconstrained real-valued kernel $K$ is  replaced by
a suitable member of the \tnkernels as described below.
If  $K \in \mathcal{R}^{l\times l\times m}$ denotes the unconstrained convolution kernel, then let
\begin{align} \label{Eq:replacement}
(f^*,\rho^*,\sigma_1^*,\sigma_2^*,B^*) = \arg \min_{f,\rho,\sigma_1,\sigma_2,B \in [0,1]^{l\times l \times m}} \left\| K - \sym{f,\rho,\sigma_1,\sigma_2,B} \right\|_2.
\end{align}
Subsequently, the kernel $K$ is replaced by $\sym{f^*,\rho^*,\sigma_1^*,\sigma_2^*,B^*}$ in the network. 
This is done for all the kernels in a layer simultaneously, and the parameters $f^*,\rho^*, \sigma_1^*,\sigma_2^*$ are frozen for the remainder of the training.
Notice however that the elements of $B^*$ are real over $[0,1]$ at this stage of the training. 
Backpropagation is used to further train and binarize $B^*$ in the ensuing period.
We used an Expectation Maximization algorithm \cite{dempster1977maximum,wu1983convergence,neal1998view} to speed-up the search in \eqref{Eq:replacement}.
When all the weights of a layer are converted to \tnkernels, the network is trained for a at least one epoch before the next layer weights are replaced.

%
\begin{figure}[hht]
\begin{center}
\scalebox{0.6}{\includegraphics{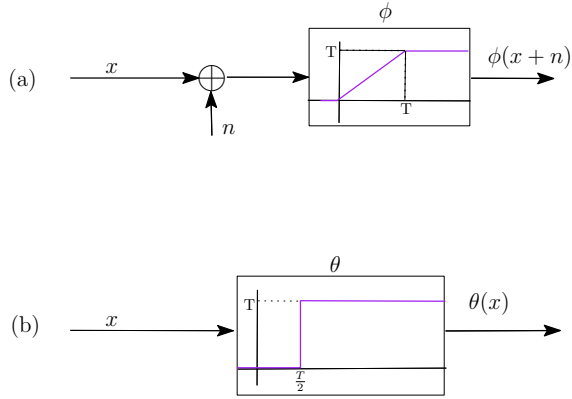}}
\end{center}
\caption{(a) Noisy ReLU, (b) Threshold neuron.}
\label{Fig:noisyReLU}
\end{figure}
\subsection{Neurons} 
Rectified linear unites (ReLUs) are the typical form of non-linearity used in deep networks.
Our approach seeks to use a binary activations obtained by threshold functions because they map to neuromorphic harware with spike-based communication infrastruture efficiently.
Our strategy is to begin the training with ReLU activations and to make them increasingly noisy so that they can be replaced with binary activations in the end.
Using stochastic neuron activation functions has been explored in the literature for various purposes \cite{williams1992simple,fiete2006gradient,noisyActivations,nair2010rectified}.
An early attempt at using noisy ReLU with the objective of obtaining binary activations was reported in \cite{wilson1994backpropagation} for single-layer networks.

It is well-known in information theory \cite{Smith2,Oettli} that the best way to communicate using an amplitude-bounded  channel in the presence of additive noise of the form shown in Figure~\ref{Fig:noisyReLU} 
is to use a discrete set of input values for $x$, and that for an appropriate choice of noise variance, binary signaling is optimal. 
Moreover, unlike in information theory where the noise distribution and variance  are presupposed, we have the freedom to choose and vary them throughout the training period.
We seek to utilize this connection to train a binary convolutional network using noisy ReLUs.

During training, we use the bounded noisy ReLU shown in Figure~\ref{Fig:noisyReLU} (a) with a zero mean uniform distributed random variable distributed
in the range $[-\epsilon,\epsilon]$ as the noise source. 
The range $\epsilon$ of the random variable is slowly increased from $0$ to $T/2$. Towards the end of the training period, noisy ReLUs are replaced by the threshold neuron shown in 
Figure~\ref{Fig:noisyReLU} (b) one layer at a time for fine tuning the network. We use the gradient of the ReLU saturating at $T$ during the backward step throughout the training.
\subsection{Systems and methods}
The networks are trained on GPUs using MatConvNet deep learning libraries \cite{matconvnet}.
The stochastic gradient descent was  used with dropout  \cite{hinton2012improving} layers, momentum ($0.9$), weight decay ($10^{-6}$), and batch normalization
\cite{batchNorm}.
The  parameters  learned  through  training  are  mapped  to hardware  using  reusable,  composable  network  description
functions  called corelets \cite{corelet}.
The  corelets  created  for  this work automatically compile the learned network parameters,
which are independent of any neuromorphic platform, into an platform-specific hardware configuration file that can directly
program TrueNorth chips.

\section{Results} \label{Sec:res}

\begin{figure}[htbp]
\begin{center}
\begin{minipage}[h]{5.2in}
\includegraphics[width=0.5in]{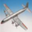}\hspace{-1pt}
\includegraphics[width=0.5in]{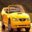}\hspace{-1pt}
\includegraphics[width=0.5in]{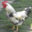}\hspace{-1pt}
\includegraphics[width=0.5in]{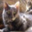}\hspace{-1pt}
\includegraphics[width=0.5in]{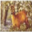}\hspace{-1pt}
\includegraphics[width=0.5in]{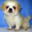}\hspace{-1pt}
\includegraphics[width=0.5in]{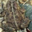}\hspace{-1pt}
\includegraphics[width=0.5in]{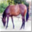}\hspace{-1pt}
\includegraphics[width=0.5in]{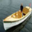}\hspace{-1pt}
\includegraphics[width=0.5in]{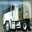}\hspace{-1pt}
\end{minipage}
\vspace{5pt}
\begin{minipage}[h]{5.2in}
\includegraphics[width=0.5in]{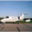}\hspace{-1pt}
\includegraphics[width=0.5in]{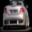}\hspace{-1pt}
\includegraphics[width=0.5in]{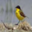}\hspace{-1pt}
\includegraphics[width=0.5in]{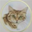}\hspace{-1pt}
\includegraphics[width=0.5in]{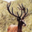}\hspace{-1pt}
\includegraphics[width=0.5in]{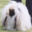}\hspace{-1pt}
\includegraphics[width=0.5in]{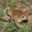}\hspace{-1pt}
\includegraphics[width=0.5in]{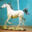}\hspace{-1pt}
\includegraphics[width=0.5in]{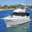}\hspace{-1pt}
\includegraphics[width=0.5in]{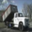}\hspace{-1pt}
\end{minipage}
\end{center}
\caption{CIFAR10 Dataset samples from 10 different classes.}
\label{Fig:Dataset}
\end{figure}

The proposed method was evaluated  using  the image recognition  benchmark  dataset CIFAR-10 (Figure~\ref{Fig:Dataset}). 
The CIFAR-10 dataset ~\cite{Krizhevsky09} consists of color natural images,  32 x 32 pixels in size,  in 10 classes, with 50000 training images and 10000 test images.

\subsection{Networks} We evaluated the representation power of \tnkernels by designing two convolution networks using one and four TrueNorth chips and trained for CIFAR10  dataset. Recent results \cite{binaryNet2,XNOR, Steve} as well as our own experiments seem to suggest that pooling layers are unsuitable for networks with binary neuron activations. 
So our networks are built by stacking four sets of convolution layers and each
set contains four convolution layers.
The smaller network is described in Table~\ref{Tab:NetworkModels} while the large network we used was obtained simply by increasing the number of features in each layer. 
The final set of output features are divided uniformly among the classes and the evaluation is performed at 1 classification per hardware tick.

\renewcommand{\tabcolsep}{1.5pt}

\begin{table}[htbp]
\centering
\begin{small}
\begin{tabular}{|c||c|c|c|c|c|c|c|c|}
\hline
 & \multicolumn{2}{ |c| }{Layer 1} & \multicolumn{2}{ |c| }{Layer 2}  & \multicolumn{2}{ |c| }{Layer 3}  & \multicolumn{2}{ |c| }{Layer 4} \\
\cline{2-9}
Layer & patch size/ & output size/ &  patch size/ & output size/ & patch size/ & output size/ &  patch size/ & output size/ \\
Sets  & stride & groups & stride & groups & stride & groups & stride & groups\\
\hline
\hline
Set 1 & 3\!$\times$\!3/1 & 32\!$\times$\!32\!$\times$\!16/1 & 3\!$\times$\!3/1 & 32\!$\times$\!32\!$\times$\!128/1 & 1\!$\times$\!1/1 & 32\!$\times$\!32\!$\times$\!128/1 & 2\!$\times$\!2/2 & 16\!$\times$\!16\!$\times$\!140/4 \\
\hline
Set 2 & 3\!$\times$\!3/1 & 16\!$\times$\!16\!$\times$\!240/20 & 1\!$\times$\!1/1 & 16\!$\times$\!16\!$\times$\!256/1 & 1\!$\times$\!1/1 & 16\!$\times$\!16\!$\times$\!256/1 &  2\!$\times$\!2/2 & 8\!$\times$\!8\!$\times$\!224/8 \\
\hline
Set 3 & 3\!$\times$\!3/1 & 8\!$\times$\!8\!$\times$\!512/32 & 1\!$\times$\!1/1 & 8\!$\times$\!8\!$\times$\!512/2 & 1\!$\times$\!1/1 & 8\!$\times$\!8\!$\times$\!512/2 & 2\!$\times$\!2/2 & 4\!$\times$\!4\!$\times$\!1024/16 \\
\hline
Set 4 & 3\!$\times$\!3/1 & 4\!$\times$\!4\!$\times$\!1024/64 & 1\!$\times$\!1/1 & 4\!$\times$\!4\!$\times$\!1024/4 & 1\!$\times$\!1/1 & 4\!$\times$\!4\!$\times$\!1024/4 & 1\!$\times$\!1/1 & 4\!$\times$\!4\!$\times$\!1000/4 \\
\hline
\end{tabular}
\end{small}
\vspace{5pt}
\caption{Structure of 1-chip symmetric kernel convolution network. The network consists of four sets and each set contains four convolution layers with varying patch size, stride and groups.}
\label{Tab:NetworkModels}
\end{table}

\subsection{Performance}
To characterize performance, the trained 1-Chip convolution networks were deployed on a IBM TrueNorth  NS1e board packed with 4096 neuromorphic cores and
run to measure the classification accuracy and throughput.
The 4-Chip networks were run on simulation \cite{Preissl12},  and classification accuracies were measured for the dataset. The results are shown in Table~\ref{Tab:Results}  in comparison with Energy-efficient deep networks (EEDN) on TrueNorth \cite{Steve}.

\renewcommand{\tabcolsep}{2pt}
\begin{table}[htbp]
\centering
\begin{small}
\begin{tabular}{|c|c c|c c|c c c|c c c|}
\hline
 State of & \multicolumn{4}{ |c| }{Multi-Chip Networks}  &\multicolumn{6}{ |c| }{1-Chip Networks} \\
\cline{2-11}
 the Art &\multicolumn{2}{ |c| }{EEDN \cite{Steve}} & \multicolumn{2}{ |c| }{Symmetric Kernel}  &\multicolumn{3}{ |c| }{EEDN} & \multicolumn{3}{ |c| }{Symmetric Kernel}\\
 Accuracy & Accuracy & \#Cores & Accuracy & \#Cores  & Accuracy & \#Cores & FPS  & Accuracy & \#Cores & FPS \\
\hline
\hline
&&&&&&&&&&\\
 91.73\% & 87.50\% & 31872 & 87.7\% &  13216 & 82.50\% & 3978 & 1191 & 84.68\% & 4044 & 934 \\
&&&&&&&&&&\\
\hline
\end{tabular}
\end{small}
\vspace{5pt}
\caption{Classification accuracy and throughput of TrueNorth Networks. FPS = Frames/Sec.}
\label{Tab:Results}
\end{table}

By using symmetric kernels, we have been able to deploy a network with twice as many features on the same number of cores as was possible using the approach described in \cite{Steve}. Thus, we have been able to obtain better accuracy for the same amount of hardware even though the set of allowed convolutional kernels are a strict subset of the set of $\{-1,0,1\}$-valued kernels allowed in \cite{Steve}. With multi-chip networks, we achieved near state-of-the-art accuracy with significantly less number of TrueNorth cores, improving the energy efficiency by more than two-fold.
 

\section{Conclusions} \label{Sec:con}
Our study shows that convolutional networks built using only structured kernels are surprisingly rich and maintain their representational power, 
and that backpropagation can be adapted to learn in the presence of such constraints.
Furthermore, our investigation suggests that 
co-designing algorithms and architecture 
may offer a successful strategy toward even more energy efficient deployment platform 
for deep learning.

\section{Appendix}
\supplementaMaterial{
\begin{remark} \label{Re:proof}
Suppose that a TrueNorth core receives a vectorized $16 \times 16$ matrix $X$ along its input-lines, and computes convolution with the Laplacian operator $K$ in Example~\ref{Ex:1}. Hence, $n=16$, $l = 3$ and it is easily verified that $W(K)$ is a $256 \times (n-l+1)^2$  (i.e., $256 \times 196$) matrix.
Since $K$ satisfies the conditions of Theorem~\ref{Th:toeplitz}, we now use $f$, $\rho$, $\sigma_1$ and $\sigma_2$ from the example 
to construct $g$, $C$ and $s_1,\ldots,s_{196}$ such that $M(g,C,\{s_i\}_{i=1}^{196}) = W(K)$.
Define the $16 \times 16$ matrix $G$ by $G_{i,j} = \sigma_1^{i-1} ( \sigma_2^{j-1} (\rho))$, thus\footnote{Reader may notice that $G$ defined here is Toeplitz. This is merely a consequence of the fact that $\sigma_1 = \sigma_2$ in this example. In general $G$ need not be Toeplitz.}
\begin{align} \label{Eq:defG}
G = \begin{bmatrix}
\rho & \sigma_2(\rho) & \cdots & \sigma_2^{15}(\rho) \\
\sigma_1(\rho) & \sigma_1(\sigma_2(\rho)) & \cdots & \sigma_1(\sigma_2^{15}(\rho)) \\
\vdots & \vdots & \vdots & \vdots \\
\sigma_1^{15}(\rho) & \sigma_1^{15}(\sigma_2(\rho)) & \cdots & \sigma_1^{15}(\sigma_2^{15}(\rho))
\end{bmatrix} = \begin{bmatrix}
1 & 2 & 1 & \cdots & 2 \\
2 & 1 & 2 & \cdots & 1 \\
1 & 2 &  1 & \cdots & 2 \\
\vdots & \vdots & \vdots & \vdots \\
2 & 1 & 2 & \cdots & 1
\end{bmatrix}
\end{align}
We use $\vect(G)$ as the type-vector $g$ in \eqref{Eq:TNweightMatrix}. 
Consequently, input element $X_{i,j}$ is assigned the type $G_{i,j}$.
Next we construct the $256 \times 196$ binary matrix $C$ one column at a time and define the strength functions.
Let  $k,l \in \{1,2,\ldots,14\}$,  let $r = (l-1)\cdot 14+k$, and define 
a $16 \times 16$ intermediate matrix $H$ by
\begin{align} \label{Eq:H}
H_{i,j} = \begin{cases}
B_{i-k+1,j-l+1} & \mbox{if $k \le i \le (k+2)$ and $l \le j \le (l+2)$} \\
0 & \mbox{otherwise}.
\end{cases}
\end{align}
Now define $C_r$ (the $r$-th column of $C$) to be $\vect(H)$, and the $r$-th strength function to be
\begin{align} \label{Eq:s}
s_{r} = f \cdot (\sigma_1^{k-1})^{-1} \cdot (\sigma_2^{l-1})^{-1}.
\end{align}
By iterating over all pairs $k$ and $l$, the matrix $C$  and all the strength functions are completely defined.
Let $k,l \in \{1,2,\ldots,14\}$,  and $r = (l-1)\cdot 14+k$.
For $g$ and $C$ and $s_1,\ldots,s_{196}$ defined above,  we have
\begin{align}
& \left( \vect(X)^t \; M(g,C,\{s_i\}_{i=1}^{196}) \right)_{r} \nonumber \\
 & \quad = \vect(X)^t \; M(g,C,\{s_i\}_{i=1}^{196})_{r} \nonumber \\
			 & \quad \overset{(a)}{=} \vect(X)^t \left(s_{r}(g) \circ C_{r}\right) \nonumber \\
			  & \quad \overset{(b)}{=} \begin{bmatrix} 1 & 1 & 1 \end{bmatrix} \left(
							\begin{bmatrix}
							X_{k,l} & X_{k,l+1} & X_{k,l+2} \\
							X_{k+1,l} & X_{k+1,l+1} & X_{k+1,l+2} \\
							X_{k+2,l} & X_{k+2,l+1} & X_{k+2,l+2} \\
							\end{bmatrix} \circ 
							\begin{bmatrix}
							0 & s_r(G_{k,l+1}) & 0 \\
							s_r(G_{k+1,l}) & s_r(G_{k+1,l+1}) & s_r(G_{k+1,l+2}) \\
							0 & s_r(G_{k+2,l+1}) & 0\\
							\end{bmatrix} \right)  \begin{bmatrix} 1 \\ 1 \\ 1\end{bmatrix} \label{Eq:TNnrn}
\end{align}
where $(a)$ follows since $M(g,C,\{s_i\}_{i=1}^{196})_{r}$ denotes the $r$-th column of $M(g,C,\{s_i\}_{i=1}^{196})$ and its $r$-th column is simply the 
element-wise product between $s_{r}(g)$ and $C_{r}$; 
$(b)$ follows from the fact that $C_r$ (defined through $H$ in \eqref{Eq:H}) masks all components of $X$ that do not contribute to the $(k,l)$-th convolution result and the 
all-one row and column vectors simply sum all elements of the $3 \times 3$ matrix between them.

Now using \eqref{Eq:s} and $G_{i,j} = \sigma_1^{i-1} (\sigma_2^{j-1} (\rho))$, we get
\begin{align}
							\begin{bmatrix}
							0 & s_r(G_{k,l+1}) & 0 \\
							s_r(G_{k+1,l}) & s_r(G_{k+1,l+1}) & s_r(G_{k+1,l+2}) \\
							0 & s_r(G_{k+2,l+1}) & 0\\
							\end{bmatrix} & = 
							\begin{bmatrix}
							0 & f(\sigma_2(\rho)) & 0 \\
							f(\sigma_1(\rho)) & f(\sigma_1(\sigma_2(\rho))) & f(\sigma_1(\sigma_2^2(\rho))) \\
							f(\sigma_1^2(\rho)) & f(\sigma_1^2(\sigma_2(\rho))) & f(\sigma_1^2(\sigma_2^2(\rho)))
							\end{bmatrix} \nonumber \\
							& \overset{(a)}{=} \begin{bmatrix}
							0 & f(2) & 0 \\
							f(2) & f(1) & f(2) \\
							0 & f(2) & 0 
							\end{bmatrix} \nonumber \\
							& \overset{(b)}{=} \begin{bmatrix}
							0 & -1 & 0 \\
							-1 & 4 & -1 \\
							0 & -1 & 0
							\end{bmatrix} \nonumber \\
							& = K \label{Eq:effWeight}
\end{align}
where $(a)$ follows by using $\rho = 1$ and applying $\sigma_1$ and $\sigma_2$ from the example,
$(b)$ is obtained by applying $f$ defined in the example.
Substituting \eqref{Eq:effWeight} in \eqref{Eq:TNnrn}, we get
\begin{align}
 \left( \vect(X)^t \; M(g,C,\{s_i\}_{i=1}^{196}) \right)_{r}  & = \begin{bmatrix} 1 & 1 & 1 \end{bmatrix} \left(
							\begin{bmatrix}
							X_{k,l} & X_{k,l+1} & X_{k,l+2} \\
							X_{k+1,l} & X_{k+1,l+1} & X_{k+1,l+2} \\
							X_{k+2,l} & X_{k+2,l+1} & X_{k+2,l+2} \\
							\end{bmatrix} \circ K\right)
							\begin{bmatrix} 1 \\ 1 \\ 1\end{bmatrix} \nonumber \\
							& \overset{(a)}{=} \left(X \coAsterisk K \right)_{k,l} \nonumber  \\
							& \overset{(b)}{=} \left( \vect(X)^t \; W(K) \right)_{r} \label{Eq:rthEquality} \\
 \vect(X)^t \; M(g,C,\{s_i\}_{i=1}^{196}) &  \overset{(c)}{=} 	\vect(X)^t \; W(K) \label{Eq:equality} \\	
 		M(g,C,\{s_i\}_{i=1}^{196})  & \overset{(d)}{=} W(K) \nonumber			
\end{align}
where $(a)$ follows from the definition of convolution, 
$(b)$ follows since $r = (l-1)\cdot 14 +k $, and $r$-th element of the row-vector $\vect(X)^t \; W(K) $ corresponds to $(k,l)$-th convolution result,
$(c)$ follows since \eqref{Eq:rthEquality} holds for each $r \in \{1,2,\ldots,196\}$, 
$(d)$ follows since $X$ is arbitrary in \eqref{Eq:equality}.
The last equality completes the proof of first part of the theorem applied to Example~\ref{Ex:1}.
\end{remark}
The following example shows the construction of the input-type vector $g$, the binary connectivity matrix $C$, and and strength functions in \eqref{Eq:TNweightMatrix} in much 
more detail for concreteness.
\begin{example} \label{Ex:toyExample}
Consider convolving a $4 \times 4$ matrix $X$ with a $3 \times 3$ kernels $K$, and denote the resulting $2 \times 2$ matrix by $Y$.
The kernel is placed on top of the corresponding elements of matrix $X$ needed to compute $Y$ below:
\begin{center}
\scalebox{0.9}{\includegraphics{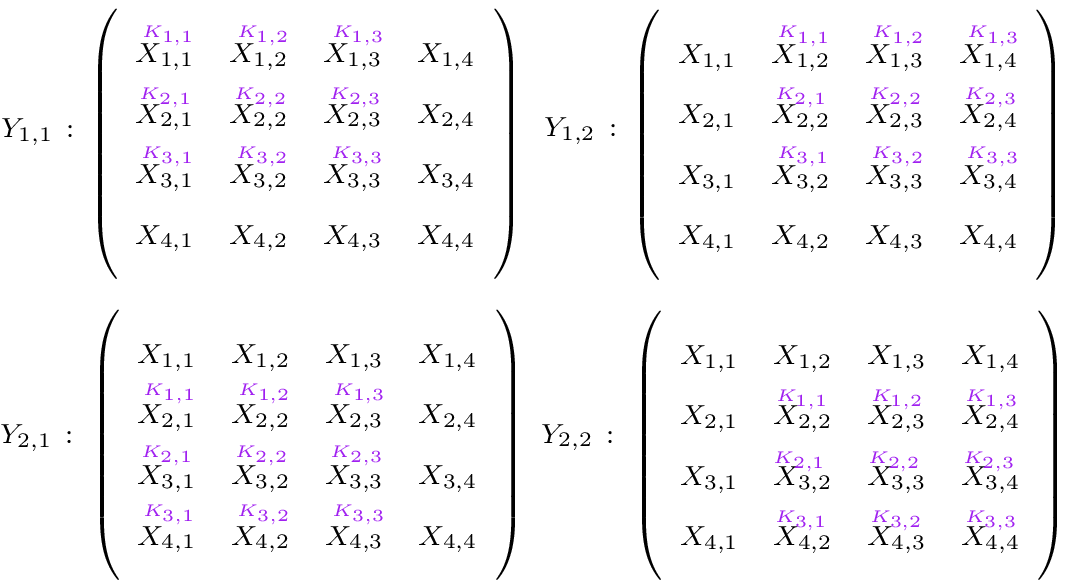}}
\end{center}
\begin{minipage}{\textwidth}
This can be expressed as matrix multiplication in the following form: 
\begin{center}
\scalebox{1.1}{\includegraphics{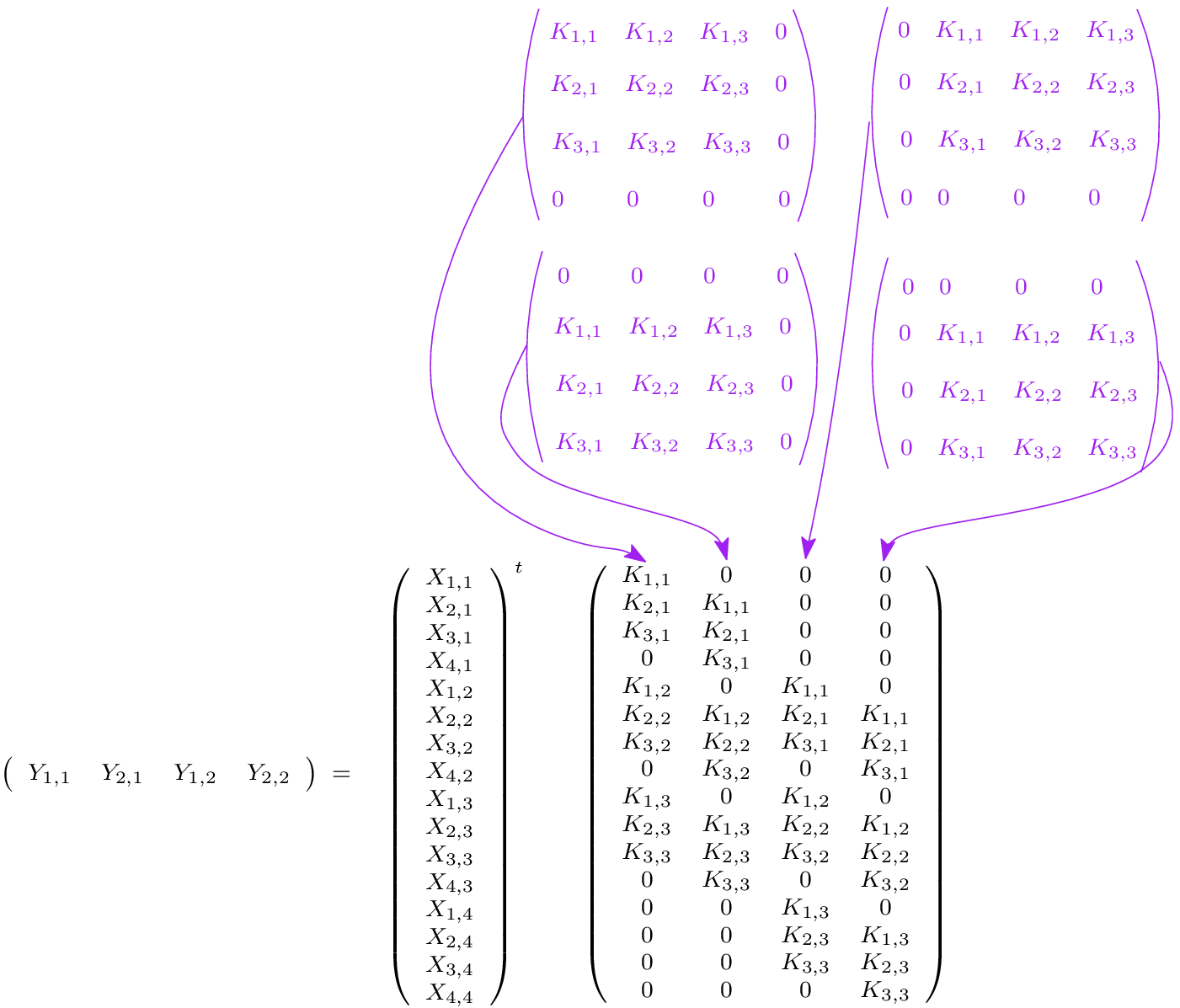}}
\end{center}
\begin{align} \label{Eq:convolutionInMatrixFormForTN}
&  = \begin{pmatrix} 
 X_{1,1} \\  X_{2,1} \\  X_{3,1} \\  X_{4,1} \\  X_{1,2} \\ X_{2,2} \\  X_{3,2} \\ X_{4,2} \\ X_{1,3} \\ X_{2,3} \\  X_{3,3} \\ X_{4,3} \\X_{1,4} \\ X_{2,4} \\  X_{3,4} \\ X_{4,4}   
 \end{pmatrix}^t \quad \left( \begin{pmatrix}
					K_{1,1} & 0 & 0 & 0 \\
					K_{2,1} & K_{1,1} & 0 & 0 \\
					K_{3,1} & K_{2,1} & 0 & 0 \\
					0 & K_{3,1} & 0 & 0\\
					K_{1,2} & 0 & K_{1,1} & 0 \\
					K_{2,2} & K_{1,2} & K_{2,1} & K_{1,1} \\
					K_{3,2} & K_{2,2} & K_{3,1} & K_{2,1} \\
					0 & K_{3,2} & 0 & K_{3,1} \\
					K_{1,3} & 0 & K_{1,2} & 0 \\
					K_{2,3} & K_{1,3} & K_{2,2} & K_{1,2}  \\
					K_{3,3} & K_{2,3} & K_{3,2} & K_{2,2} \\
					0 & K_{3,3} & 0 & K_{3,2} \\
					0 & 0 & K_{1,3} & 0 \\
					0 & 0 & K_{2,3} & K_{1,3} \\
					0 & 0 & K_{3,3} & K_{2,3} \\
					0 & 0 & 0 & K_{3,3}          
					\end{pmatrix} \circ \begin{bmatrix}
					1 & 0 & 0 & 0 \\
					1 & 1 & 0 & 0 \\
					1 & 1 & 0 & 0 \\
					0 & 1 & 0 & 0 \\
					1 & 0 & 1 & 0 \\
					1 & 1 & 1 & 1 \\					
					1 & 1 & 1 & 1 \\
					0 & 1 & 0 & 1 \\
					1 & 0	 & 1 & 0 \\	
					1 & 1 & 1 & 1 \\
					1 & 1 & 1 & 1 \\
					0 & 1 & 0 & 1 \\
					0 & 0 & 1 & 0 \\
					0 & 0 & 1 & 1 \\
					0 & 0 & 1 & 1 \\
					0 & 0	 & 0 & 1
					\end{bmatrix} \right)
\end{align} 
\end{minipage}
Now consider performing the above matrix multiplication using a TrueNorth core. 
Naturally, the input vector $\vect(X)$ is routed to $16$ of the input lines of a TrueNorth core in the order shown in \eqref{Eq:convolutionInMatrixFormForTN}, 
and the two matrices on either side of the Hadamard product in \eqref{Eq:convolutionInMatrixFormForTN} are implemented using the crossbar by selecting
$1)$ an appropriate binary connectivity matrix $C$, $2)$ a set of four strength functions $s_1, s_2, s_3$, and $s_4$, and $3)$ a $4 \times 4$ matrix $G$ such that
$G_{i,j}$ defines the input-type for $X_{i,j}$ so that by defining $g = \vect(G)$ we can get
\begin{align} \label{Eq:TNmatrixCondition0}
 [ s_1(g) \; s_2(g) \; s_3(g) \; s_4(g)] \circ C  & = 
					\begin{pmatrix}
					K_{1,1} & 0 & 0 & 0 \\
					K_{2,1} & K_{1,1} & 0 & 0 \\
					K_{3,1} & K_{2,1} & 0 & 0 \\
					0 & K_{3,1} & 0 & 0\\
					K_{1,2} & 0 & K_{1,1} & 0 \\
					K_{2,2} & K_{1,2} & K_{2,1} & K_{1,1} \\
					K_{3,2} & K_{2,2} & K_{3,1} & K_{2,1} \\
					0 & K_{3,2} & 0 & K_{3,1} \\
					K_{1,3} & 0 & K_{1,2} & 0 \\
					K_{2,3} & K_{1,3} & K_{2,2} & K_{1,2}  \\
					K_{3,3} & K_{2,3} & K_{3,2} & K_{2,2} \\
					0 & K_{3,3} & 0 & K_{3,2} \\
					0 & 0 & K_{1,3} & 0 \\
					0 & 0 & K_{2,3} & K_{1,3} \\
					0 & 0 & K_{3,3} & K_{2,3} \\
					0 & 0 & 0 & K_{3,3}          
					\end{pmatrix} \circ \begin{bmatrix}
					1 & 0 & 0 & 0 \\
					1 & 1 & 0 & 0 \\
					1 & 1 & 0 & 0 \\
					0 & 1 & 0 & 0 \\
					1 & 0 & 1 & 0 \\
					1 & 1 & 1 & 1 \\					
					1 & 1 & 1 & 1 \\
					0 & 1 & 0 & 1 \\
					1 & 0	 & 1 & 0 \\	
					1 & 1 & 1 & 1 \\
					1 & 1 & 1 & 1 \\
					0 & 1 & 0 & 1 \\
					0 & 0 & 1 & 0 \\
					0 & 0 & 1 & 1 \\
					0 & 0 & 1 & 1 \\
					0 & 0	 & 0 & 1
					\end{bmatrix}
\end{align}
which is sufficient to be able to compute the aforementioned convolution.

The obvious choice for $C$ is the binary matrix on the RHS of \eqref{Eq:TNmatrixCondition0}, and the remainder of the discussion focuses on the choice of the type vector and strength functions.
Since the strength function is common to all the rows of a particular column, two rows with distinct non-zero entries in a column must be assigned distinct types. 
To illustrate this further, let us consider the following kernel
\begin{align*}
K = \begin{pmatrix}
-1 & 2 & -1 \\
-2 & 4 & -2 \\
-1 & 2 & -1
\end{pmatrix}.
\end{align*}
Now let us rewrite the matrix on the RHS of \eqref{Eq:TNmatrixCondition0} for this example and color the rows of the resulting matrix such that any two rows of the first column with distinct non-zero elements have different colors:
\begin{align*} 
\begin{pmatrix}
					K_{1,1} & 0 & 0 & 0 \\
					K_{2,1} & K_{1,1} & 0 & 0 \\
					K_{3,1} & K_{2,1} & 0 & 0 \\
					0 & K_{3,1} & 0 & 0\\
					K_{1,2} & 0 & K_{1,1} & 0 \\
					K_{2,2} & K_{1,2} & K_{2,1} & K_{1,1} \\
					K_{3,2} & K_{2,2} & K_{3,1} & K_{2,1} \\
					0 & K_{3,2} & 0 & K_{3,1} \\
					K_{1,3} & 0 & K_{1,2} & 0 \\
					K_{2,3} & K_{1,3} & K_{2,2} & K_{1,2}  \\
					K_{3,3} & K_{2,3} & K_{3,2} & K_{2,2} \\
					0 & K_{3,3} & 0 & K_{3,2} \\
					0 & 0 & K_{1,3} & 0 \\
					0 & 0 & K_{2,3} & K_{1,3} \\
					0 & 0 & K_{3,3} & K_{2,3} \\
					0 & 0 & 0 & K_{3,3}          
					\end{pmatrix} & = \begin{pmatrix}
					\rowcolor{green!10} 
					 -1 & 0 & 0 & 0 \\
					 \rowcolor{red!20}  
					-2 & -1 & 0 & 0 \\
					\rowcolor{green!10}
					-1 & -2 & 0 & 0 \\
					0 & -1 & 0 & 0\\
					\rowcolor{blue!20}
					2 & 0 & -1 & 0 \\
					\rowcolor{yellow!20}
					4 & 2 & -2 & -1 \\
					\rowcolor{blue!20}
					2 & 4 & -1 & -2 \\
					0 & 2 & 0 & -1 \\
					\rowcolor{green!10}
					-1 & 0 & 2 & 0 \\
					\rowcolor{red!20}
					-2 & -1 & 4 & 2  \\
					\rowcolor{green!10}
					-1 & -2 & 2 & 4 \\
					0 & -1 & 0 & 2 \\
					0 & 0 & -1 & 0 \\
					0 & 0 & -2 & -1 \\
					0 & 0 & -1 & -2 \\
					0 & 0 & 0 & -1         
					\end{pmatrix}
\end{align*}
The rows where the first column is $0$ are free to choose a color based on subsequent columns so they remain uncolored for now.
However, even though the colors so far have been chosen considering only the first column, it is important that no two rows of a subsequent column get the same color but contain distinct non-zero entries. 
In that case, the coloring scheme is said to be in {\it `conflict'}, and a conflict-free coloring scheme may not exist. 
Notice that the current row coloring does not contain any such conflicts so we can proceed with the remaining columns one at a time (starting position of the row-color indicates
which column was under consideration while a row was assigned a particular color):
\begin{minipage}{\textwidth}
\begin{align*}
 \begin{pmatrix}
					\rowcolor{green!10} 
					 -1 & 0 & 0 & 0 \\
					 \rowcolor{red!20}  
					-2 & -1 & 0 & 0 \\
					\rowcolor{green!10}
					-1 & -2 & 0 & 0 \\
					0 & \tTwo -1 & \tTwo 0 & \tTwo 0\\
					\rowcolor{blue!20} 
					2 & 0 & -1 & 0 \\
					\rowcolor{yellow!20} 
					4 & 2 & -2 & -1 \\
					\rowcolor{blue!20}
					2 & 4 & -1 & -2 \\
					0 & \tFou 2 & \tFou 0 & \tFou -1 \\
					\rowcolor{green!10}
					-1 & 0 & 2 & 0 \\
					\rowcolor{red!20}
					-2 & -1 & 4 & 2  \\
					\rowcolor{green!10}
					-1 & -2 & 2 & 4 \\
					0 & \tTwo -1 & \tTwo 0 & \tTwo 2 \\
					0 & 0 & \tThr -1 & \tThr 0 \\
					0 & 0 & \tFou  -2 & \tFou -1 \\
					0 & 0 & \tThr -1 & \tThr -2 \\
					0 & 0 & 0 & \tFou -1         
					\end{pmatrix} \implies g = \begin{pmatrix} 
					1 \\
					2 \\
					1 \\
					2 \\
					3\\
					4\\
					3\\
					4\\
					1\\
					2\\
					1\\
					2\\
					3\\
					4\\
					3\\
					4
					\end{pmatrix}  \quad \left. \begin{matrix} 
					(1,1) \\
					(2,1) \\
					(3,1) \\
					(4,1) \\
					(1,2) \\
					(2,2) \\
					(3,2) \\
					(4,2) \\
					(1,3) \\
					(2,3) \\
					(3,3) \\
					(4,3) \\
					(1,4) \\
					(2,4) \\
					(3,4) \\
					(4,4)
					\end{matrix} \right\} \overset{\overset{\mbox{$(i,j)$ location of the input incident}}{\xleftarrow{\hspace*{5.2cm}}}}{\mbox{on the corresponding row as in \eqref{Eq:convolutionInMatrixFormForTN} }} 
\end{align*}
\vspace{-0.6cm}
\begin{center}
\hspace{-1.25cm}
\scalebox{1}{\includegraphics{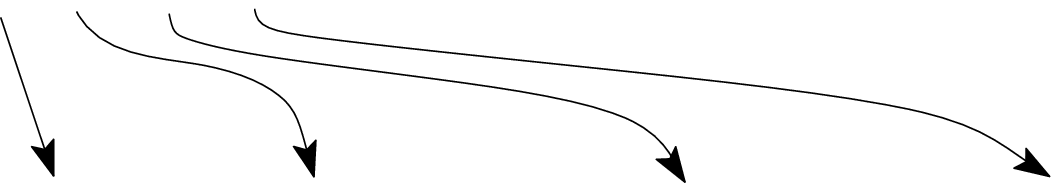}}
\end{center}
\vspace{-0.5cm}
\begin{align*}
\begin{pmatrix}
\tOne -1 & \tThr 2 & \tOne -1 & \tThr 0 \\
\tTwo -2 & \tFou 4 & \tTwo -2 & \tFou 0 \\
\tOne -1 & \tThr 2 & \tOne -1 & \tThr 0 \\
\tTwo 0 & \tFou 0 & \tTwo 0 & \tFou 0
\end{pmatrix} \quad
\begin{pmatrix}
\tOne 0 & \tThr 0 & \tOne 0 & \tThr 0 \\
\tTwo -1 & \tFou 2 & \tTwo -1 & \tFou 0 \\
\tOne -2 & \tThr 4 & \tOne -2 & \tThr 0 \\
\tTwo -1 & \tFou 2 & \tTwo -1 & \tFou 0 
\end{pmatrix} \quad
\begin{pmatrix}
\tOne 0 & \tThr -1 & \tOne 2 & \tThr -1  \\
\tTwo 0 & \tFou -2 & \tTwo 4 & \tFou -2 \\
\tOne 0 & \tThr -1 & \tOne 2 & \tThr -1 \\
\tTwo 0 & \tFou 0 & \tTwo 0 & \tFou 0 
\end{pmatrix} \quad
\begin{pmatrix}
\tOne 0 & \tThr 0 & \tOne 0 & \tThr 0 \\
\tTwo 0 & \tFou -1 & \tTwo 2 & \tFou -1  \\
\tOne 0 & \tThr -2 & \tOne 4 & \tThr -2  \\
\tTwo 0 & \tFou -1 & \tTwo 2 & \tFou  -1 
\end{pmatrix}  \\
\begin{matrix}
s_1(1) & : & -1 \\
s_1(2) & : & -2 \\
s_1(3) & : & 2 \\
s_1(4) & : & 4 
\end{matrix} \quad \quad \quad \quad
\begin{matrix}
s_2(1) & : & -2 \\
s_2(2) & : & -1 \\
s_2(3) & : & 4 \\
s_2(4) & : & 2 
\end{matrix} \quad \quad \quad \quad
\begin{matrix}
s_3(1) & : & 2 \\
s_3(2) & : & 4 \\
s_3(3) & : & -1 \\
s_3(4) & : & -2 
\end{matrix} \quad \quad \quad \quad
\begin{matrix}
s_4(1) & : & 4 \\
s_4(2) & : & 2 \\
s_4(3) & : & -2 \\
s_4(4) & : & -1 
\end{matrix} 
\end{align*}
\end{minipage}
where the type-vector $g$ simply indicates the color index of the corresponding row (green -- 1, red -- 2, blue -- 3, yellow -- 4). The above coloring scheme represents a valid coloring solution that is conflict-free.
For $i = 1,2,3,4$, the strength function $s_i$ is defined by letting $s_i(c)$ to be the unique non-zero value associated with color index $c$ in column $i$ and we have a solution that
satisfies \eqref{Eq:TNmatrixCondition0}.

In general, suppose that the kernel $K$ satisfies $K_{i,j} = f(\sigma_1^{i-1}(\sigma_2^{j-1}(\rho)))$ for some function $f$, $\rho \in \{1,2,3,4\}$, and two commuting permutations $\sigma_1$ 
and $\sigma_2$ on $\{1,2,3,4\}$.
We will now illustrate  the idea behind why  the above coloring procedure will never run into conflicts for such a $K$.
For simplicity, assume that the function $f$ is invertible.
Since such a $K$ has exactly four distinct entries, rows of the matrix whose first column contains a non-zero entry can be colored with four distinct colors without any conflict in that column.
Assume that after such a coloring based on just the non-zero entries of the first column, two rows have been already assigned the same color as shown below:
\begin{align*}
\begin{pmatrix}
					K_{1,1} & 0 & 0 & 0 \\
					K_{2,1} & K_{1,1} & 0 & 0 \\
					K_{3,1} & K_{2,1} & 0 & 0 \\
					0 & K_{3,1} & 0 & 0\\
					K_{1,2} & 0 & K_{1,1} & 0 \\
					\rowcolor{green!10}
					K_{2,2} & K_{1,2} & K_{2,1} & K_{1,1} \\
					K_{3,2} & K_{2,2} & K_{3,1} & K_{2,1} \\
					0 & K_{3,2} & 0 & K_{3,1} \\
					K_{1,3} & 0 & K_{1,2} & 0 \\
					K_{2,3} & K_{1,3} & K_{2,2} & K_{1,2}  \\
					\rowcolor{green!10}
					K_{3,3} & K_{2,3} & K_{3,2} & K_{2,2} \\
					0 & K_{3,3} & 0 & K_{3,2} \\
					0 & 0 & K_{1,3} & 0 \\
					0 & 0 & K_{2,3} & K_{1,3} \\
					0 & 0 & K_{3,3} & K_{2,3} \\
					0 & 0 & 0 & K_{3,3}          
					\end{pmatrix}
\end{align*}
 The two rows with the same color imply
 \begin{align*}
 K_{2,2} & = K_{3,3} \\
 \implies f(\sigma_1(\sigma_2(\rho)))  & = f(\sigma_1^2(\sigma_2^2(\rho))) \\
 \implies \sigma_1(\sigma_2(\rho)) & = \sigma_1^2(\sigma_2^2(\rho)) \\
 \implies \rho & = \sigma_1(\sigma_2(\rho)).
 \end{align*}
 Now let us use the last equality to show that this coloring choice does not lead to a conflict in any of the three remaining columns:
 \begin{align*}
 K_{2,3} & = \sigma_1(\sigma_2^2(\rho))  = \sigma_2(\sigma_1(\sigma_2(\rho)))  = \sigma_2(\rho)  = K_{1,2} \\
 K_{3,2} & = \sigma_1^2(\sigma_2(\rho))  =\sigma_1(\sigma_1(\sigma_2(\rho))) =  \sigma_1(\rho)  = K_{2,1} \\
 K_{2,2} & = \sigma_1(\sigma_2(\rho))  = \rho  = K_{1,1}
 \end{align*}
 which is precisely what we needed.
 Similar analysis can be used to show that all the rows can be colored without conflicts for such a matrix $K$.
 In fact, by defining the $4 \times 4$ matrix $G$ by $G_{i,j} = \sigma_1^{i-1}(\sigma_2^{j-1}(\rho))$ and letting $g = \vect(G)$, we can use 
 $g$ to obtain a conflict-free coloring solution!
 \end{example}
\begin{proof}[Outline of the proof of Theorem~\ref{Th:toeplitz}]
Extending the arguments in Remark~\ref{Re:proof} in a straightforward manner proves the first part of the theorem.
We will now outline the proof for the second half the theorem assuming $n=16$,  $l =3$ and the proof for general case is a straightforward extension of arguments here. 
Assume that $K$ contains at least four distinct non-zero entries as in the statement of the theorem.
Suppose that there exist some $g$, $C$, and $\{s_i\}_{i=1}^{N}$ such that $M(g,C,\{s_i\}_{i=1}^{N}) = W(K)$,
we will now construct $f$, $\rho$, $\sigma_1$ , and $\sigma_2$ such that 
$$
K_{i,j} = f(\sigma_1^{i-1} (\sigma_2^{j-1} (\rho))).
$$
Let $k,l \in \{1,2,\ldots, 14\}$, and $r = (l-1)\cdot 14+k$.
Let $X$ denote the $16 \times 16$ input matrix.
As in Remark~\ref{Re:proof}, the $(k,l)$-th convolution result is computed using the $r$-th column of $M(g,C,\{s_i\}_{i=1}^{N}) $ by 
\begin{align} \label{Eq:converseTNconv}
\vect(X)^t \, M(g,C,\{s_i\}_{i=1}^{N})_r & = \vect(X)^t \,  \left(s_r(g) \circ C_r\right).
\end{align}
Recall that the range of $s_r$ cannot contain zero (since its range must contain all unique entries of K -- of which there are four non-zero elements), 
consequently $s_r(g)$ cannot not contain any zero entries.
The $(k,l)$-th convolution only depends on $\{X_{i,j} : k \le i \le (k+2), l \le j \le (l+2)\}$.
Furthermore, since the theorem statement also requires that $K$ does not contain zeros, the $(k,l)$-th convolution depends on all of $\{X_{i,j} : k \le i \le (k+2), l \le j \le (l+2)\}$
non-trivially.
Hence, if $H$ denotes the $16 \times 16$ matrix constructed such that $C_r = \vect(H)$), then $H$ must satisfy
\begin{align} \label{Eq:Hdef}
H_{i,j} = \begin{cases}
1 & \mbox{if $k \le i \le (k+2)$ and $l \le j \le (l+2)$} \\
0 & \mbox{otherwise}
\end{cases}
\end{align}
which is merely stating that the binary vector $C_r$ masks out all the rows on which the $(k,l)$-th convolution does not depend on.
Now let  $G$ denote the $16 \times 16$ matrix obtained by reshaping $g$ so that the input-type of $X_{i,j}$ is $G_{i,j}$ (i.e, $g = \vect(G)$).
We have
\begin{align}
& \vect(X)^t \,  \left(s_r(g) \circ C_r\right) \nonumber \\
& \quad \overset{(a)}{=} \begin{bmatrix} 1 & 1 & 1 \end{bmatrix} \left(
							\begin{bmatrix}
							X_{k,l} & X_{k,l+1} & X_{k,l+2} \\
							X_{k+1,l} & X_{k+1,l+1} & X_{k+1,l+2} \\
							X_{k+2,l} & X_{k+2,l+1} & X_{k+2,l+2} 
							\end{bmatrix} \circ s_r \left(
							\begin{bmatrix}
							G_{k,l} & G_{k,l+1} & G_{k,l+2} \\
							G_{k+1,l} & G_{k+1,l+1} & G_{k+1,l+2} \\
							G_{k+2,l} & G_{k+2,l+1} & G_{k+2,l+2}
							\end{bmatrix} \right) \right)  \begin{bmatrix} 1 \\ 1 \\ 1\end{bmatrix} \label{Eq:TNconv3} \\
& \quad \overset{(b)}{=} \begin{bmatrix} 1 & 1 & 1 \end{bmatrix} \left(
							\begin{bmatrix}
							X_{k,l} & X_{k,l+1} & X_{k,l+2} \\
							X_{k+1,l} & X_{k+1,l+1} & X_{k+1,l+2} \\
							X_{k+2,l} & X_{k+2,l+1} & X_{k+2,l+2} 
							\end{bmatrix} \circ K \right)  \begin{bmatrix} 1 \\ 1 \\ 1\end{bmatrix} \label{Eq:TNconv4}							
\end{align}
where $(a)$ follows from \eqref{Eq:Hdef} and the fact that $g = \vect(G)$,
$(b)$ follows from the assumption that $\vect(X)^t \,  \left(s_r(g) \circ C_r\right) $ is equal to the $(k,l)$-th convolution result.
Since the input matrix $X$ is arbitrary, by combining \eqref{Eq:TNconv3} and \eqref{Eq:TNconv4} we get
\begin{align} \label{Eq:KfromCol_r}
K & = s_r \left(
		\begin{bmatrix}
		G_{k,l} & G_{k,l+1} & G_{k,l+2} \\
		G_{k+1,l} & G_{k+1,l+1} & G_{k+1,l+2} \\
		G_{k+2,l} & G_{k+2,l+1} & G_{k+2,l+2}
		\end{bmatrix} \right).   
\end{align}
%
Finally, we have
\begin{align} 
s_1 \left(
\begin{bmatrix}
G_{1,1} & G_{1,2} & G_{1,3} \\
G_{2,1} & G_{2,2} & G_{2,3} \\
G_{3,1}  & G_{3,2} & G_{3,3}\\
\end{bmatrix} \right) & \overset{(a)}{=} s_{2} \left(
\begin{bmatrix}
G_{2,1} & G_{2,2} & G_{2,3} \\
G_{3,1} & G_{3,2} & G_{3,3} \\
G_{4,1}  & G_{4,2} & G_{4,3}\\
\end{bmatrix} 
\right) \nonumber \\
s_2^{-1} \cdot s_1 \left(\begin{bmatrix}
G_{1,1} & G_{1,2} & G_{1,3} \\
G_{2,1} & G_{2,2} & G_{2,3} \\
G_{3,1}  & G_{3,2} & G_{3,3}\\
\end{bmatrix} \right) & \overset{(b)}{=} 
\begin{bmatrix}
G_{2,1} & G_{2,2} & G_{2,3} \\
G_{3,1} & G_{3,2} & G_{3,3} \\
G_{4,1}  & G_{4,2} & G_{4,3}\\
\end{bmatrix} \label{Eq:sigma1} \\
s_1 \left(
\begin{bmatrix}
G_{1,1} & G_{1,2} & G_{1,3} \\
G_{2,1} & G_{2,2} & G_{2,3} \\
G_{3,1}  & G_{3,2} & G_{3,3}\\
\end{bmatrix} \right)
& \overset{(c)}{=}  s_{15} \left(
\begin{bmatrix}
G_{1,2} & G_{1,3} & G_{1,4} \\
G_{2,2} & G_{2,3} & G_{2,4} \\
G_{3,2}  & G_{3,3} & G_{3,4}\\
\end{bmatrix} \right) \nonumber \\
s_{15}^{-1} \cdot s_1 \left(
\begin{bmatrix}
G_{1,1} & G_{1,2} & G_{1,3} \\
G_{2,1} & G_{2,2} & G_{2,3} \\
G_{3,1}  & G_{3,2} & G_{3,3}\\
\end{bmatrix} \right)
& = 
\begin{bmatrix}
G_{1,2} & G_{1,3} & G_{1,4} \\
G_{2,2} & G_{2,3} & G_{2,4} \\
G_{3,2}  & G_{3,3} & G_{3,4}\\
\end{bmatrix}  \label{Eq:sigma2}
\end{align}
where $(a)$ follows from the fact that \eqref{Eq:KfromCol_r} is valid for all $r$ and by equating right-hand-side of \eqref{Eq:KfromCol_r} for $(k,l) = (1,1)$ and $(k,l) = (2,1)$,
$(b)$ follows from observing that if $K$ contains four distinct non-zero elements the domain and range of the strength functions contain exactly four elements, thus they become bijections and hence inverses exist, 
and
$(c)$ is similar to $(a)$ but using $(k,l)=(1,1)$ and $(k,l) = (1,2)$.

By defining $f = s_1$, $\sigma_1 = s_2^{-1} \cdot s_1$,  $\sigma_2 = s_{15}^{-1} \cdot s_1$, and $\rho = G_{1,1}$ we get the desired construction as follows
\begin{align*}
K & \overset{(a)}{=} f \left( 
\begin{bmatrix}
G_{1,1} & G_{1,2} & G_{1,3} \\
G_{2,1} & G_{2,2} & G_{2,3} \\
G_{3,1}  & G_{3,2} & G_{3,3}\\
\end{bmatrix} 
\right)  \\
& \overset{(b)}{=} f \left(\begin{bmatrix}
G_{1,1} & \sigma_2 (G_{1,1}) & \sigma_2^2(G_{1,1}) \\
\sigma_1(G_{1,1}) & \sigma_1 (\sigma_2(G_{1,1})) &  \sigma_1 (\sigma_2^2 (G_{1,1})) \\
\sigma_1^2(G_{1,1})  &  \sigma_1^2( \sigma_2(G_{1,1})) &  \sigma_1^2( \sigma_2^2(G_{1,1}))\\
\end{bmatrix} \right) \\
& \overset{(c)}{=} f \left(\begin{bmatrix}
G_{1,1} & \sigma_2 (G_{1,1}) & \sigma_2^2(G_{1,1}) \\
\sigma_1(G_{1,1}) & \sigma_2 (\sigma_1(G_{1,1})) &  \sigma_2^2 (\sigma_1 (G_{1,1})) \\
\sigma_1^2(G_{1,1})  &  \sigma_2( \sigma_1^2(G_{1,1})) &  \sigma_2^2( \sigma_1^2(G_{1,1}))\\
\end{bmatrix} \right) \\
\end{align*}
where $(a)$ follows from using \eqref{Eq:KfromCol_r} with $(k,l) = (1,1)$ and $f=s_1$,
$(b)$ follows from \eqref{Eq:sigma1} and \eqref{Eq:sigma2} (for example, \eqref{Eq:sigma1} implies  $G_{2,2} = \sigma_1(G_{1,2})$ while \eqref{Eq:sigma2} 
implies $G_{1,2} = \sigma_2(G_{1,1})$, so $G_{2,2} = \sigma_1(G_{1,2}) = \sigma_1(\sigma_2(G_{1,1}))$),
and $(c)$ follows by applying  \eqref{Eq:sigma1} and \eqref{Eq:sigma2} but the order reversed from before--for example, \eqref{Eq:sigma2} implies $G_{2,2} = \sigma_2(G_{2,1})$
and \eqref{Eq:sigma1} implies $G_{2,1} = \sigma_1(G_{1,1})$, hence $G_{2,2} = \sigma_2(G_{2,1}) = \sigma_2( \sigma_1(G_{1,1}))$.
The last two equalities show $\sigma_1$ and $\sigma_2$ commute on elements of $K$. Since $K$ contains four distinct elements (which is exactly the number of elements in the dommain of $\sigma_1$ and $\sigma_2$), they must commute for the sake of uniqueness of $K$.
\end{proof}
}
\small
\bibliography{TNkernelsSparseNets} 

\begin{thebibliography}{10}
\providecommand{\url}[1]{#1}
\csname url@samestyle\endcsname
\providecommand{\newblock}{\relax}
\providecommand{\bibinfo}[2]{#2}
\providecommand{\BIBentrySTDinterwordspacing}{\spaceskip=0pt\relax}
\providecommand{\BIBentryALTinterwordstretchfactor}{4}
\providecommand{\BIBentryALTinterwordspacing}{\spaceskip=\fontdimen2\font plus
\BIBentryALTinterwordstretchfactor\fontdimen3\font minus
  \fontdimen4\font\relax}
\providecommand{\BIBforeignlanguage}[2]{{%
\expandafter\ifx\csname l@#1\endcsname\relax
\typeout{** WARNING: IEEEtran.bst: No hyphenation pattern has been}%
\typeout{** loaded for the language `#1'. Using the pattern for}%
\typeout{** the default language instead.}%
\else
\language=\csname l@#1\endcsname
\fi
#2}}
\providecommand{\BIBdecl}{\relax}
\BIBdecl

\bibitem{alexNet}
A.~Krizhevsky, I.~Sutskever, and G.~E. Hinton, ``Imagenet classification with
  deep convolutional neural networks,'' in \emph{Advances in neural information
  processing systems}, 2012, pp. 1097--1105.

\bibitem{silver2016mastering}
D.~Silver, A.~Huang, C.~J. Maddison, A.~Guez, L.~Sifre, G.~Van Den~Driessche,
  J.~Schrittwieser, I.~Antonoglou, V.~Panneershelvam, M.~Lanctot \emph{et~al.},
  ``Mastering the game of go with deep neural networks and tree search,''
  \emph{Nature}, vol. 529, no. 7587, pp. 484--489, 2016.

\bibitem{Steve}
\BIBentryALTinterwordspacing
S.~K. Esser, P.~A. Merolla, J.~V. Arthur, A.~S. Cassidy, R.~Appuswamy,
  A.~Andreopoulos, D.~J. Berg, J.~L. McKinstry, T.~Melano, D.~R. Barch,
  C.~di~Nolfo, P.~Datta, A.~Amir, B.~Taba, M.~D. Flickner, and D.~S. Modha,
  ``Convolutional networks for fast, energy-efficient neuromorphic computing,''
  \emph{CoRR}, vol. abs/1603.08270, 2016. [Online]. Available:
  \url{http://arxiv.org/abs/1603.08270}
\BIBentrySTDinterwordspacing

\bibitem{binaryNet}
\BIBentryALTinterwordspacing
M.~Courbariaux and Y.~Bengio, ``Binarynet: Training deep neural networks with
  weights and activations constrained to +1 or -1,'' \emph{CoRR}, vol.
  abs/1602.02830, 2016. [Online]. Available:
  \url{http://arxiv.org/abs/1602.02830}
\BIBentrySTDinterwordspacing

\bibitem{binaryNet2}
\BIBentryALTinterwordspacing
I.~Hubara, D.~Soudry, and R.~E. Yaniv, ``Binarized neural networks,''
  \emph{CoRR}, vol. abs/1602.02505, 2016. [Online]. Available:
  \url{http://arxiv.org/abs/1602.02505}
\BIBentrySTDinterwordspacing

\bibitem{binaryConnect}
M.~Courbariaux, Y.~Bengio, and J.-P. David, ``Binaryconnect: Training deep
  neural networks with binary weights during propagations,'' in \emph{Advances
  in Neural Information Processing Systems}, 2015, pp. 3105--3113.

\bibitem{compressMobile}
\BIBentryALTinterwordspacing
Y.~Kim, E.~Park, S.~Yoo, T.~Choi, L.~Yang, and D.~Shin, ``Compression of deep
  convolutional neural networks for fast and low power mobile applications,''
  \emph{CoRR}, vol. abs/1511.06530, 2015. [Online]. Available:
  \url{http://arxiv.org/abs/1511.06530}
\BIBentrySTDinterwordspacing

\bibitem{XNOR}
\BIBentryALTinterwordspacing
M.~Rastegari, V.~Ordonez, J.~Redmon, and A.~Farhadi, ``Xnor-net: Imagenet
  classification using binary convolutional neural networks,'' \emph{CoRR},
  vol. abs/1603.05279, 2016. [Online]. Available:
  \url{http://arxiv.org/abs/1603.05279}
\BIBentrySTDinterwordspacing

\bibitem{Merolla}
P.~A. Merolla, J.~V. Arthur, R.~Alvarez-Icaza, A.~S. Cassidy, J.~Sawada,
  F.~Akopyan, B.~L. Jackson, N.~Imam, C.~Guo, Y.~Nakamura, B.~Brezzo, I.~Vo,
  S.~K. Esser, R.~Appuswamy, B.~Taba, A.~Amir, M.~D. Flickner, W.~P. Risk,
  R.~Manohar, and D.~S. Modha, ``A million spiking-neuron integrated circuit
  with a scalable communication network and interface,'' \emph{Science}, vol.
  345, no. 6197, pp. 668--673, 2014.

\bibitem{smallFootprint}
\BIBentryALTinterwordspacing
V.~Sindhwani, T.~N. Sainath, and S.~Kumar, ``Structured transforms for
  small-footprint deep learning,'' in \emph{Neural Information Processing
  Systems (NIPS)}, 2015. [Online]. Available:
  \url{http://arxiv.org/pdf/1510.01722v1.pdf}
\BIBentrySTDinterwordspacing

\bibitem{Moczulski2015}
M.~Moczulski, M.~Denil, J.~Appleyard, and N.~de~Freitas, ``Acdc: A structured
  efficient linear layer,'' in \emph{International Conference on Learning
  Representations (ICLR)}, 2016.

\bibitem{ionescu2015matrix}
C.~Ionescu, O.~Vantzos, and C.~Sminchisescu, ``Matrix backpropagation for deep
  networks with structured layers,'' in \emph{Proceedings of the IEEE
  International Conference on Computer Vision}, 2015, pp. 2965--2973.

\bibitem{rankConstrainedTop}
P.~Nakkiran, R.~Alvarez, R.~Prabhavalkar, and C.~Parada, ``Compressing deep
  neural networks using a rank-constrained topology,'' in \emph{Proceedings of
  Annual Conference of the International Speech Communication Association
  (Interspeech)}, 2015, pp. 1473--1477.

\bibitem{vecQuant}
\BIBentryALTinterwordspacing
Y.~Gong, L.~Liu, M.~Yang, and L.~D. Bourdev, ``Compressing deep convolutional
  networks using vector quantization,'' \emph{CoRR}, vol. abs/1412.6115, 2014.
  [Online]. Available: \url{http://arxiv.org/abs/1412.6115}
\BIBentrySTDinterwordspacing

\bibitem{deepComp}
\BIBentryALTinterwordspacing
S.~Han, H.~Mao, and W.~J. Dally, ``Deep compression: Compressing deep neural
  network with pruning, trained quantization and huffman coding,'' \emph{CoRR},
  vol. abs/1510.00149, 2015. [Online]. Available:
  \url{http://arxiv.org/abs/1510.00149}
\BIBentrySTDinterwordspacing

\bibitem{hashNet}
\BIBentryALTinterwordspacing
W.~Chen, J.~T. Wilson, S.~Tyree, K.~Q. Weinberger, and Y.~Chen, ``Compressing
  neural networks with the hashing trick,'' \emph{CoRR}, vol. abs/1504.04788,
  2015. [Online]. Available: \url{http://arxiv.org/abs/1504.04788}
\BIBentrySTDinterwordspacing

\bibitem{optimalBrain}
Y.~L. Cun, J.~S. Denker, and S.~A. Solla, ``Optimal brain damage,'' in
  \emph{Advances in Neural Information Processing Systems}.\hskip 1em plus
  0.5em minus 0.4em\relax Morgan Kaufmann, 1990, pp. 598--605.

\bibitem{bottleneckFeaturesTara}
\BIBentryALTinterwordspacing
T.~N. Sainath, B.~Kingsbury, and B.~Ramabhadran, ``Auto-encoder bottleneck
  features using deep belief networks,'' in \emph{2012 {IEEE} International
  Conference on Acoustics, Speech and Signal Processing, {ICASSP} 2012, Kyoto,
  Japan, March 25-30, 2012}, 2012, pp. 4153--4156. [Online]. Available:
  \url{http://dx.doi.org/10.1109/ICASSP.2012.6288833}
\BIBentrySTDinterwordspacing

\bibitem{bottleneckFeaturesPetr}
F.~Gr{\'e}zl and P.~Fousek, ``Optimizing bottle-neck features for lvcsr,'' in
  \emph{Acoustics, Speech and Signal Processing, 2008. ICASSP 2008. IEEE
  International Conference on}.\hskip 1em plus 0.5em minus 0.4em\relax IEEE,
  2008, pp. 4729--4732.

\bibitem{lowRankMatrixTara}
T.~N. Sainath, B.~Kingsbury, V.~Sindhwani, E.~Arisoy, and B.~Ramabhadran,
  ``Low-rank matrix factorization for deep neural network training with
  high-dimensional output targets,'' in \emph{Acoustics, Speech and Signal
  Processing (ICASSP), 2013 IEEE International Conference on}.\hskip 1em plus
  0.5em minus 0.4em\relax IEEE, 2013, pp. 6655--6659.

\bibitem{EIE}
\BIBentryALTinterwordspacing
S.~Han, X.~Liu, H.~Mao, J.~Pu, A.~Pedram, M.~A. Horowitz, and W.~J. Dally,
  ``{EIE:} efficient inference engine on compressed deep neural network,''
  \emph{CoRR}, vol. abs/1602.01528, 2016. [Online]. Available:
  \url{http://arxiv.org/abs/1602.01528}
\BIBentrySTDinterwordspacing

\bibitem{spinnaker}
E.~Painkras, L.~A. Plana, J.~Garside, S.~Temple, F.~Galluppi, C.~Patterson,
  D.~R. Lester, A.~D. Brown, and S.~B. Furber, ``Spinnaker: A 1-w 18-core
  system-on-chip for massively-parallel neural network simulation,''
  \emph{Solid-State Circuits, IEEE Journal of}, vol.~48, no.~8, pp. 1943--1953,
  2013.

\bibitem{pfeil2012six}
T.~Pfeil, A.~Gr{\"u}bl, S.~Jeltsch, E.~M{\"u}ller, P.~M{\"u}ller, M.~A.
  Petrovici, M.~Schmuker, D.~Br{\"u}derle, J.~Schemmel, and K.~Meier, ``Six
  networks on a universal neuromorphic computing substrate,'' \emph{arXiv
  preprint arXiv:1210.7083}, 2012.

\bibitem{schmuker2014neuromorphic}
M.~Schmuker, T.~Pfeil, and M.~P. Nawrot, ``A neuromorphic network for generic
  multivariate data classification,'' \emph{Proceedings of the National Academy
  of Sciences}, vol. 111, no.~6, pp. 2081--2086, 2014.

\bibitem{moradi2014event}
S.~Moradi and G.~Indiveri, ``An event-based neural network architecture with an
  asynchronous programmable synaptic memory,'' \emph{Biomedical Circuits and
  Systems, IEEE Transactions on}, vol.~8, no.~1, pp. 98--107, 2014.

\bibitem{park201465k}
J.~Park, S.~Ha, T.~Yu, E.~Neftci, and G.~Cauwenberghs, ``A 65k-neuron
  73-mevents/s 22-pj/event asynchronous micro-pipelined integrate-and-fire
  array transceiver,'' in \emph{Biomedical Circuits and Systems Conference
  (BioCAS), 2014 IEEE}.\hskip 1em plus 0.5em minus 0.4em\relax IEEE, 2014, pp.
  675--678.

\bibitem{frigo2005design}
M.~Frigo and S.~G. Johnson, ``The design and implementation of fftw3,''
  \emph{Proceedings of the IEEE}, vol.~93, no.~2, pp. 216--231, 2005.

\bibitem{maass2004computational}
W.~Maass and H.~Markram, ``On the computational power of circuits of spiking
  neurons,'' \emph{Journal of computer and system sciences}, vol.~69, no.~4,
  pp. 593--616, 2004.

\bibitem{dongarra2000guest}
J.~Dongarra and F.~Sullivan, ``Guest editors? introduction: The top 10
  algorithms,'' \emph{Computing in Science \& Engineering}, vol.~2, no.~1, pp.
  22--23, 2000.

\bibitem{hoare1962quicksort}
C.~A. Hoare, ``Quicksort,'' \emph{The Computer Journal}, vol.~5, no.~1, pp.
  10--16, 1962.

\bibitem{bottou2010large}
L.~Bottou, ``Large-scale machine learning with stochastic gradient descent,''
  in \emph{Proceedings of COMPSTAT'2010}.\hskip 1em plus 0.5em minus
  0.4em\relax Springer, 2010, pp. 177--186.

\bibitem{binary1}
\BIBentryALTinterwordspacing
Z.~Wu, D.~Lin, and X.~Tang, ``Adjustable bounded rectifiers: Towards deep
  binary representations,'' \emph{http://arxiv.org/abs/1511.06201}, vol.
  abs/1511.06201, 2015. [Online]. Available:
  \url{http://arxiv.org/abs/1511.06201}
\BIBentrySTDinterwordspacing

\bibitem{neuron}
A.~S. Cassidy, P.~Merolla, J.~V. Arthur, S.~K. Esser, B.~Jackson,
  R.~Alvarez-Icaza, P.~Datta, J.~Sawada, T.~M. Wong, V.~Feldman \emph{et~al.},
  ``Cognitive computing building block: A versatile and efficient digital
  neuron model for neurosynaptic cores,'' in \emph{Neural Networks (IJCNN), The
  2013 International Joint Conference on}.\hskip 1em plus 0.5em minus
  0.4em\relax IEEE, 2013, pp. 1--10.

\bibitem{gray2006toeplitz}
R.~M. Gray, \emph{Toeplitz and circulant matrices: A review}.\hskip 1em plus
  0.5em minus 0.4em\relax now publishers inc, 2006.

\bibitem{van2014optimizing}
B.~Van~Werkhoven, J.~Maassen, H.~E. Bal, and F.~J. Seinstra, ``Optimizing
  convolution operations on gpus using adaptive tiling,'' \emph{Future
  Generation Computer Systems}, vol.~30, pp. 14--26, 2014.

\bibitem{cuDNN}
\BIBentryALTinterwordspacing
S.~Chetlur, C.~Woolley, P.~Vandermersch, J.~Cohen, J.~Tran, B.~Catanzaro, and
  E.~Shelhamer, ``cudnn: Efficient primitives for deep learning,'' \emph{CoRR},
  vol. abs/1410.0759, 2014. [Online]. Available:
  \url{http://arxiv.org/abs/1410.0759}
\BIBentrySTDinterwordspacing

\bibitem{qadeer2013convolution}
W.~Qadeer, R.~Hameed, O.~Shacham, P.~Venkatesan, C.~Kozyrakis, and M.~A.
  Horowitz, ``Convolution engine: balancing efficiency \& flexibility in
  specialized computing,'' in \emph{ACM SIGARCH Computer Architecture News},
  vol.~41, no.~3.\hskip 1em plus 0.5em minus 0.4em\relax ACM, 2013, pp. 24--35.

\bibitem{dempster1977maximum}
A.~P. Dempster, N.~M. Laird, and D.~B. Rubin, ``Maximum likelihood from
  incomplete data via the em algorithm,'' \emph{Journal of the royal
  statistical society. Series B (methodological)}, pp. 1--38, 1977.

\bibitem{wu1983convergence}
C.~J. Wu, ``On the convergence properties of the em algorithm,'' \emph{The
  Annals of statistics}, pp. 95--103, 1983.

\bibitem{neal1998view}
R.~M. Neal and G.~E. Hinton, ``A view of the em algorithm that justifies
  incremental, sparse, and other variants,'' in \emph{Learning in graphical
  models}.\hskip 1em plus 0.5em minus 0.4em\relax Springer, 1998, pp. 355--368.

\bibitem{williams1992simple}
R.~J. Williams, ``Simple statistical gradient-following algorithms for
  connectionist reinforcement learning,'' \emph{Machine learning}, vol.~8, no.
  3-4, pp. 229--256, 1992.

\bibitem{fiete2006gradient}
I.~R. Fiete and H.~S. Seung, ``Gradient learning in spiking neural networks by
  dynamic perturbation of conductances,'' \emph{Physical review letters},
  vol.~97, no.~4, p. 048104, 2006.

\bibitem{noisyActivations}
\BIBentryALTinterwordspacing
{\c{C}}.~G{\"{u}}l{\c{c}}ehre, M.~Moczulski, M.~Denil, and Y.~Bengio, ``Noisy
  activation functions,'' \emph{CoRR}, vol. abs/1603.00391, 2016. [Online].
  Available: \url{http://arxiv.org/abs/1603.00391}
\BIBentrySTDinterwordspacing

\bibitem{nair2010rectified}
V.~Nair and G.~E. Hinton, ``Rectified linear units improve restricted boltzmann
  machines,'' in \emph{Proceedings of the 27th International Conference on
  Machine Learning (ICML-10)}, 2010, pp. 807--814.

\bibitem{wilson1994backpropagation}
E.~Wilson, ``Backpropagation learning for systems with discrete-valued
  functions,'' in \emph{Proceedings of the World Congress on Neural Networks},
  vol.~3, 1994, pp. 332--339.

\bibitem{Smith2}
J.~G. Smith, ``The information capacity of amplitude and variance constrained
  scalar gaussian channel,'' \emph{Information and Control}, vol.~18, p.
  203?219, 1971.

\bibitem{Oettli}
W.~Oettli, ``The capacity-achieving input distribution for some ampli-
  tude-limited channels with additive noise,'' \emph{IEEE Trans. Inform.
  Theory}, vol. IT?20, p. 372?374, May 1974.

\bibitem{matconvnet}
A.~Vedaldi and K.~Lenc, ``Matconvnet: Convolutional neural networks for
  matlab,'' in \emph{Proceedings of the 23rd Annual ACM Conference on
  Multimedia Conference}.\hskip 1em plus 0.5em minus 0.4em\relax ACM, 2015, pp.
  689--692.

\bibitem{hinton2012improving}
G.~E. Hinton, N.~Srivastava, A.~Krizhevsky, I.~Sutskever, and R.~R.
  Salakhutdinov, ``Improving neural networks by preventing co-adaptation of
  feature detectors,'' \emph{arXiv preprint arXiv:1207.0580}, 2012.

\bibitem{batchNorm}
S.~Ioffe and C.~Szegedy, ``Batch normalization: Accelerating deep network
  training by reducing internal covariate shift,'' in \emph{Proceedings of The
  32nd International Conference on Machine Learning}, 2015, pp. 448--456.

\bibitem{corelet}
A.~Amir, P.~Datta, W.~P. Risk, A.~S. Cassidy, J.~A. Kusnitz, S.~K. Esser,
  A.~Andreopoulos, T.~M. Wong, M.~Flickner, R.~Alvarez-Icaza \emph{et~al.},
  ``Cognitive computing programming paradigm: a corelet language for composing
  networks of neurosynaptic cores,'' in \emph{Neural Networks (IJCNN), The 2013
  International Joint Conference on}.\hskip 1em plus 0.5em minus 0.4em\relax
  IEEE, 2013, pp. 1--10.

\bibitem{Krizhevsky09}
A.~Krizhevsky, ``Learning multiple layers of features from tiny images,''
  \emph{Master's thesis, Department of Computer Science, University of
  Toronto}, 2009.

\bibitem{Preissl12}
R.~Preissl, T.~M. Wong, P.~Datta, M.~Flickner, R.~Singh, S.~K. Esser, W.~P.
  Risk, H.~D. Simon, and D.~S. Modha, ``Compass: a scalable simulator for an
  architecture for cognitive computing,'' in \emph{{SC} Conference on High
  Performance Computing Networking, Storage and Analysis}, 2012, p.~54.

\end{thebibliography}
\bibliographystyle{IEEEtran}
\end{document}